\pdfoutput=1

\documentclass[11pt,x11names]{article}
\usepackage{setspace}
\usepackage[]{misc/acl}

\usepackage{times}
\usepackage{latexsym}
\usepackage[table,x11names]{xcolor}
\usepackage{inconsolata}
\usepackage{graphicx}
\usepackage[T1]{fontenc}
\usepackage[utf8]{inputenc}
\usepackage{microtype}
\usepackage{booktabs}
\usepackage{todonotes}
\usepackage{amsmath,amssymb}
\usepackage{cleveref}
\usepackage{xspace}
\usepackage{soul} %
\usepackage{caption}
\usepackage{multirow,multicol}
\usepackage{ulem}
\usepackage{float}
\usepackage{array}
\usepackage{bm}
\usepackage{xfrac}
\usepackage{tcolorbox} %
\usepackage[shortlabels]{enumitem}
\usepackage[outline]{contour}%
\usepackage{tikz}
\usepackage{hyperref}
\usepackage{pgfplots}
\usepackage{algorithm}
\usepackage{algpseudocode}
\usepackage{siunitx}
\usepackage{twemojis}

\pgfplotsset{compat=1.18}

\setlist[itemize]{noitemsep,left=0mm}

\usepackage{tabularx}

\crefname{lstlisting}{listing}{listings}
\Crefname{lstlisting}{Listing}{Listings}
\crefname{equ}{equation}{equations}
\Crefname{equ}{Equation}{Equations}
\Crefname{algorithm}{Algorithm}{Algorithms}
\crefname{example}{example}{examples}
\Crefname{example}{Example}{Examples}
\crefname{prompt}{prompt}{prompts}
\Crefname{prompt}{Prompt}{Prompts}
\DeclareCaptionType{example}[Example][List of examples]
\DeclareCaptionType{prompt}[Prompt][List of prompts]
\DeclareCaptionType{equ}[Equation][List of equations]

\usepackage{marginnote}

\definecolor{TodoColor}{rgb}{1,0.8,0.1} %
\definecolor{TodoColor2}{rgb}{1,0.3,0.9} %
\definecolor{TodoColor3}{rgb}{0,0.9,0.6} %

\makeatletter\def\Hy@Warning#1{}\makeatother
\let\svthefootnote\thefootnote
\newcommand\blankfootnote[1]{%
  \let\thefootnote\relax\footnotetext{#1}%
  \let\thefootnote\svthefootnote%
}

\graphicspath{{img/}}

\title{
On the Limits of Model Merging for Multilinguality in Pre-Training

}

\author{
  Seth Aycock\textsuperscript{\includegraphics[height=6.85pt]{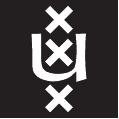}} \quad
  Fedor Vitiugin\textsuperscript{\includegraphics[height=7.7pt]{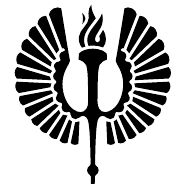}} \quad
  Aleksandr Umnov\textsuperscript{\includegraphics[height=6.85pt]{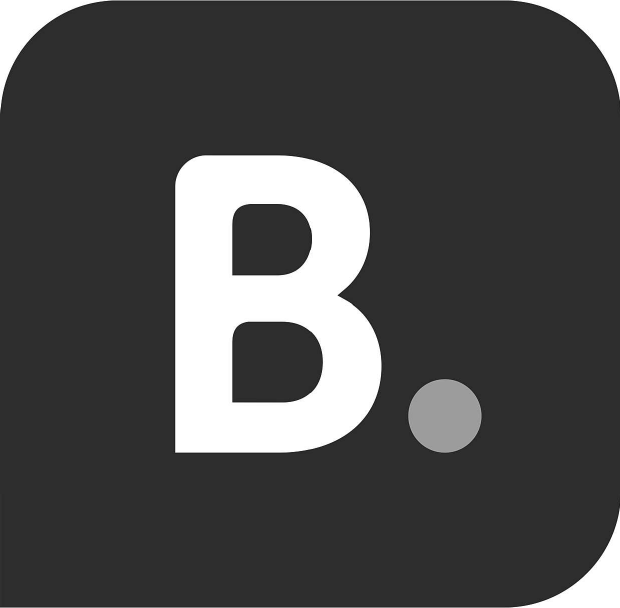}} \quad
  Christof Monz\textsuperscript{\includegraphics[height=6.85pt]{img/uva-1.pdf}} \quad
  Khalil Sima\textquotesingle an\textsuperscript{\includegraphics[height=6.85pt]{img/uva-1.pdf}} \\
  \textsuperscript{\includegraphics[height=6.85pt]{img/uva-1.pdf}}University of Amsterdam \quad
  \textsuperscript{\includegraphics[height=7.7pt]{img/turku-1.pdf}}University of Turku \quad
  \textsuperscript{\includegraphics[height=6.85pt]{img/booking-1-1.pdf}}Booking.com \\
  \texttt{s.aycock@uva.nl}
}

\begin{document}

\maketitle

\maketitle

\begin{abstract}

Endowing models with consistent multilingual performance can be achieved by \textit{mixing} pre-training data, or post-training approaches such as language-specific model \textit{merging}. In this work, we test whether merging can be applied to monolingually pre-trained models. We conduct a controlled study on the efficacy of mixed, merged, and monolingual pre-training setups. We find that while monolingual pre-training results in strong in-language performance, merging any combination of monolingual models leads to performance collapse due to interference. Our analysis suggests representational similarity is a prerequisite for model merging. We therefore conclude that the flexibility of merging in fine-tuning does not extend trivially to language-specific pre-training.

\end{abstract}

\section{Introduction }

Multilinguality is a key desideratum in training large language models (LLMs), but consistent capabilities across languages are difficult to achieve \citep{Moskvina26-MultilingualLargeLanguage}, due to both data choices~\citep{Shani26-RootsPerformanceDisparity} and modelling choices~\citep{Chang24-WhenMultilingualityCurse}. Common approaches involve mixed pre-training for early language exposure~\citep{Foroutan25-RevisitingMultilingualData, longpre2026atlas}, or fine-tuning a pre-trained model on language-specific data~\citep{aggarwal2024towards, Salamanca26-TinyAyaBridging}. These methods are performant but somewhat inflexible, requiring further adaptation to modify language coverage.

\begin{figure}[t!]
  \centering
  \hspace{-0.2cm}
  \includegraphics[width=0.495\textwidth]{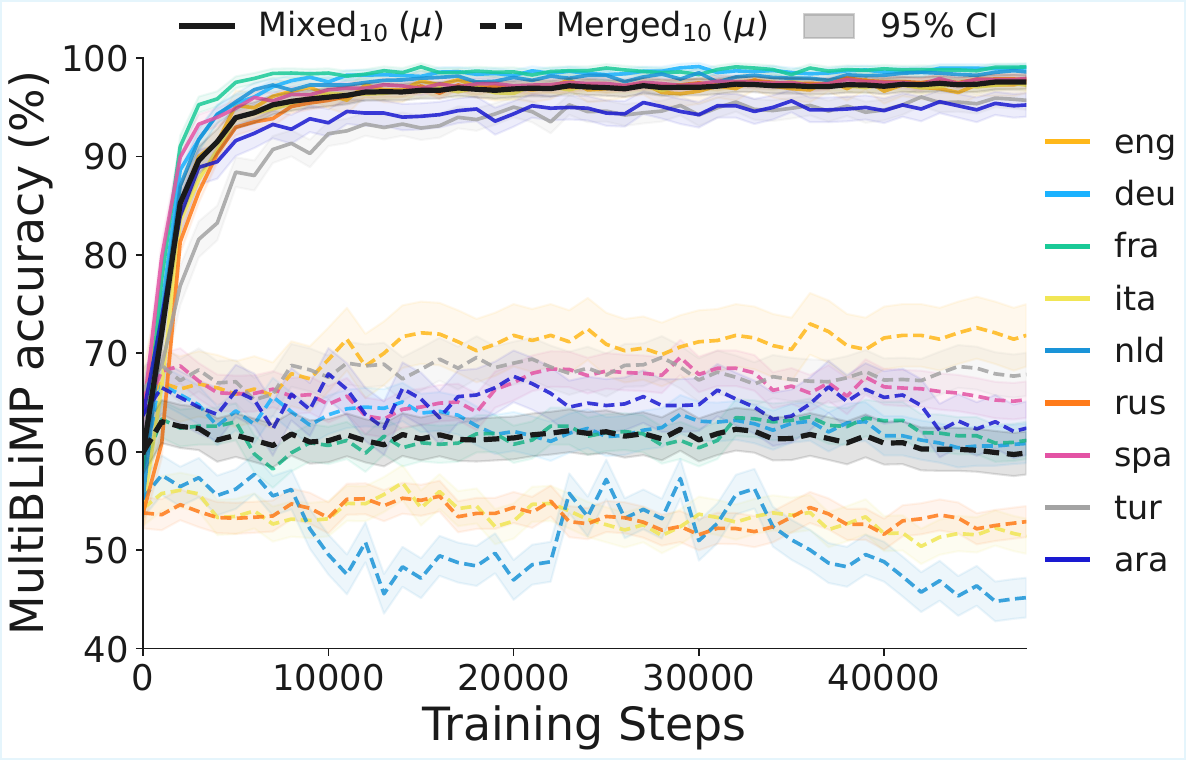}
  \caption{Average and per-language MultiBLiMP accuracies over training for a mixed data model (Mixed$_{10}$) and a linearly merged model (Merged$_{10}$) combining 10 monolingual models at each training step. Scores near 100\% are expected for grammatically competent models; 50\% indicates random chance. Equal data mixing in pre-training gives consistent multilingual performance, while merging leads to near-random performance.} %
  \label{fig:mix-vs-merge}
\end{figure}

Model merging has emerged as a cheap, post-hoc, and flexible method for improving models' multilingual or multi-task capabilities~\citep{ilharco2023editing, Bandarkar25-UnreasonableEffectivenessModela}. Standardly, task or language-specific experts are \textit{fine-tuned} from a shared \textit{pre-trained} model, then merged~\citep{Chronopoulou24-LanguageTaskArithmetic, Parovic24-InvestigatingPotentialTask, Yang24-ModelMergingLLMs, Cohere25-CommandEnterpriseReadyLarge, Zeng25-EfficientModelEditing}, optionally with parameter interference mitigating methods applied~\citep{yu2024language}. In fine-tuning settings with only a few languages or tasks, this setup can outperform data mixing~\citep{Aakanksha24-MixDataMerge, Yang25-MixDataMerge}. However, the picture for language-specific merging is less clear: recent work suggests multilingual merging for fine-tuned models can suffer from weight-space incompatibilities~\citep{Gain26-OneModelTranslate} and sometimes underperform mixed fine-tuning~\citep{Glocker25-GrowMergeScaling}. Further, while fine-tuning is cheaper than pre-training, the choice of base model constrains capabilities for all downstream evaluations. This raises the question: can merging, and its potential benefits, be extended to language-specific \textit{pre-training}?

We investigate this research question through controlled experiments: we pre-train a mixed multilingual model and test against comparable open-source monolingual models, evaluating various merging methods, and testing all models across 10 languages over 5 benchmarks. We observe that merging monolingual pre-trained models leads to a collapse in models' capabilities due to interference. Conversely, mixed data training results in consistent but modest multilingual performance. Our analysis suggests that independently pre-trained models' representations diverge too far for merging to succeed. We conclude that the flexibility of model merging does not extend trivially to independently pre-trained models despite homogenous architectures, and that some alignment is required before language-specific adaptation.

\section{Experimental Methodology}

\paragraph{Model Pre-Training}
We use the open-source HPLT 2.15B monolingual decoder-only models \citep{OpenEuroLLM25-Release38Monolingual}, which were each trained on 100B tokens of HPLT language-specific data~\citep{deGibert24-NewMassiveMultilingual, Burchell25-ExpandedMassiveMultilingual}. HPLT v2 \citep{arefyev2025hplt} is a large-scale open multilingual corpus constructed through web-crawling and language filtering.
HPLT models use the Gemma-3 tokenizer and follow the Llama architecture \citep{Touvron23-LlamaOpenFoundation} with 24 layers, 32 attention heads, and a sequence length of 2048. We pre-train a mixed data model with the same architecture (Mixed$_{10}$) on 100B tokens from 10 languages (10B tokens per language), to compare monolingual, mixed, and merged pre-training strategies. Pre-training was run with Megatron-LM \citep{Shoeybi20-MegatronLMTrainingMultiBillion}, using 16 nodes with AMD MI250x GPUs for 3,000 GPU hours on the LUMI supercomputer, for an estimated carbon footprint of 59 kg CO\textsubscript{2} per model.

While these models are relatively small and require fewer pre-training resources than closed-source alternatives, achieving state-of-the-art performance is not the primary objective of this work. Instead, we target controllability and scientific transparency. This design provides a foundation for reproducible experiments comparing different multilingual pre-training strategies, letting us systematically isolate and evaluate cross-lingual and monolingual performance across tasks\footnote{We make our \href{https://huggingface.co/collections/sethjsa/pt-merge}{models} and \href{https://github.com/Sethjsa/PT-MERGE}{code} openly available.}.

\paragraph{Model Merging}
We use linear weight averaging~\citep{Wortsman22-ModelSoupsAveraging} to equally merge the 10 monolingual HPLT models giving Merged$_{10}$ models; and we merge all 45 bilingual combinations of HPLT models for analysis in Section \ref{sec:analysis}.
We also apply the interference-mitigation method, TIES~\citep{Yadav23-TIESMergingResolvingInterference}; here, we calculate a \textit{task vector} for each trained model by subtracting the base models' parameters, which here is the shared random initialisation. After this, TIES  prunes lower magnitude parameters given a threshold, and resolves sign conflicts, then linearly adds the resulting task vectors to the base model. We also test DARE-TIES~\citep{yu2024language}, which randomly drops parameters and rescales before applying TIES. We note these methods are designed to work on fine-tunes of a base model where task vectors are of small magnitudes; however our pre-trained models see 100B tokens and we therefore expect both large magnitude differences in the task vectors calculated from the shared initialisation, and varied weight-spaces, which may compromise these methods. We merge models using Mergekit~\citep{Goddard24-ArceesMergeKitToolkit}.

\newcommand{\best}[1]{\textbf{#1}}
\newcommand{\na}{--}
\begin{table*}[t]
\centering
\footnotesize
\setlength{\tabcolsep}{5pt}

\begin{tabular}{@{} l 
    c   %
    c                      %
    S[table-format=2.1] S[table-format=2.2]   %
    S[table-format=2.1] S[table-format=2.2]   %
    S[table-format=2.1] S[table-format=2.2]   %
    S[table-format=2.1] S[table-format=2.2]   %
    S[table-format=2.1] S[table-format=2.2]   %
    @{}}
\toprule
\textbf{Model} & {\textbf{Params/B}} & {\textbf{Tokens/B}} & \multicolumn{2}{c}{MultiBLiMP} & \multicolumn{2}{c}{Belebele} & \multicolumn{2}{c}{HellaSwag} & \multicolumn{2}{c}{X-CSQA} & \multicolumn{2}{c}{FLORES} \\
\cmidrule(lr){4-5} \cmidrule(lr){6-7} \cmidrule(lr){8-9} \cmidrule(lr){10-11} \cmidrule(lr){12-13}
 & &  & {$\mu$} & {CV\%} & {$\mu$} & {CV\%} & {$\mu$} & {CV\%} & {$\mu$} & {CV\%} & {$\mu$} & {CV\%} \\
\midrule
\multicolumn{13}{l}{\textit{Multilingual baselines}} \\[2pt]
EuroLLM-1.7B$_{35}$ & 1.7 & 4000 & 95.9 & 3.62 & 36.5 & 7.34 & 45.9 & 13.44 & 31.6 & 15.65 & 37.0 & 33.89 \\
Gemma-2-2B & 2.6 & 2000 & 94.8 & 3.46 & 48.6 & 11.16 & 52.2 & 17.45 & 43.8 & 21.86 & \multicolumn{1}{r}{\textbf{42.2}} & 27.72 \\
Tiny-Aya-Base$_{70}$ & 3.35 & 6000 & 97.2 & 2.14 & \multicolumn{1}{r}{\textbf{54.3}} & 8.96 & \multicolumn{1}{r}{\textbf{56.8}} & 13.15 & \multicolumn{1}{r}{\textbf{47.8}} & 14.48 & 39.2 & 35.65 \\

\midrule
\multicolumn{13}{l}{\textit{Monolingual experts}} \\[2pt]
HPLT$_1$ ($\mu$) & 2.15 & 100* & \multicolumn{1}{r}{\textbf{98.7}} & 0.78 & 39.3 & 7.74 & 48.6 & 14.30 & 39.0 & 18.45 & 37.3 & 25.99 \\
\midrule
\multicolumn{13}{l}{\textit{Mixed pre-training}} \\[2pt]
Mixed$_{10}$ & 2.15 & 100 & 97.6 & 1.39 & 34.9 & 13.50 & 37.5 & 17.97 & 32.3 & 21.70 & 30.0 & 33.28 \\
\midrule
\multicolumn{13}{l}{\textit{Merged$_{10}$ experts}} \\[2pt]
Linear & 2.15 & 1000 & 59.9 & 14.23 & 23.9 & 6.43 & 25.1 & 6.18 & 20.3 & 7.93 & 2.1 & 17.24 \\
TIES & 2.15 & 1000 & 60.0 & 10.79 & 24.4 & 6.11 & 25.3 & 6.02 & 20.4 & 6.02 & 12.3 & 48.11 \\
DARE-TIES & 2.15 & 1000 & 60.8 & 10.63 & 24.4 & 4.40 & 25.6 & 4.53 & 20.4 & 6.83 & 1.7 & 24.67 \\
\bottomrule
\end{tabular}
\caption{%
    Mean results over 10 languages across tasks (ChrF++ for FLORES, token-normalised accuracy \% for other tasks). $\mu$ = mean performance across languages; CV = coefficient of variation (\%) measuring cross-lingual consistency. Model parameters and pre-training tokens indicated in billions (B); *HPLT$_1$ monolingual experts have seen 100B tokens \textit{each}; supported model languages, if available, in subscript. \textbf{Bold} = best result per column. Mixed pre-training gives consistent performance while merging leads to catastrophic interference and near-random results.
}
\label{tab:main-results}
\end{table*}

\paragraph{Baselines}
We test similarly-sized baselines: EuroLLM (1.7B)~\citep{Martins24-EuroLLMMultilingualLanguage} as an open-data multilingual pre-trained model, and Gemma-2 (2B)~\citep{Riviere24-Gemma2Improving} and Tiny Aya Base (3.35B)~\citep{Salamanca26-TinyAyaBridging}, with open-weights but private data mixes. These models were trained on 20-60x more tokens than HPLT models.

\paragraph{Evaluation}
We test models on 5 multilingual benchmarks: MultiBLiMP ~\citep{Jumelet26-MultiBLiMP10Massively} testing formal language competence on linguistic minimal pairs; Belebele~\citep{Bandarkar24-BelebeleBenchmarkParallel} to test reading comprehension; multilingual HellaSwag~\citep{Lai23-OkapiInstructiontunedLarge} for functional language understanding; X-CSQA~\citep{Lin21-CommonSenseEnglish} testing common-sense reasoning; and FLORES-200 translation from eng--xxx~\citep{Costa-jussa24-ScalingNeuralMachine} to test cross-lingual generation. We select 10 diverse, high-resource languages across the benchmarks: Arabic (ara), German (deu), English (eng), French (fra), Italian (ita), Dutch (nld), Russian (rus), Spanish (spa), Turkish (tur), and Mandarin Chinese (zho)\footnote{We note that MultiBLiMP lacks zho, X-CSQA lacks tur, and for FLORES in eng, we use fra--eng.}.
Belebele, HellaSwag, X-CSQA, and MultiBLiMP are evaluated with token-normalised accuracy, and run in a cloze-formulation measuring log-likelihoods; and we evaluate FLORES-200 with ChrF++~\citep{Popovic17-ChrFWordsHelping}. We calculate 95\% confidence intervals as $1.96\times \text{SE}$, estimating standard error (SE) from 1000 resamples over per-item scores.

\section{Results}

\paragraph{Monolingual models are strong oracles} Results in Table \ref{tab:main-results} show the HPLT$_{1}$ monolingual models perform best among models on test when averaged across languages, and even outperform the stronger baselines on MultiBLiMP despite substantially shorter training, establishing a high performance ceiling. This suggests monolingual pre-training is effective for training in-language competence, even for small models, aligning with prior work~\citep{Chang26-GoldfishMonolingualLanguage}. We note that scores near 100\% are expected for grammatically competent models and do not indicate overfitting \citep{Jumelet26-MultiBLiMP10Massively}. MultiBLiMP is thus a highly discriminative benchmark for our purposes, and we focus ablation analyses here accordingly.

\begin{figure}[t!]
  \centering
  \hspace{-0.2cm}
  \includegraphics[width=0.495\textwidth]{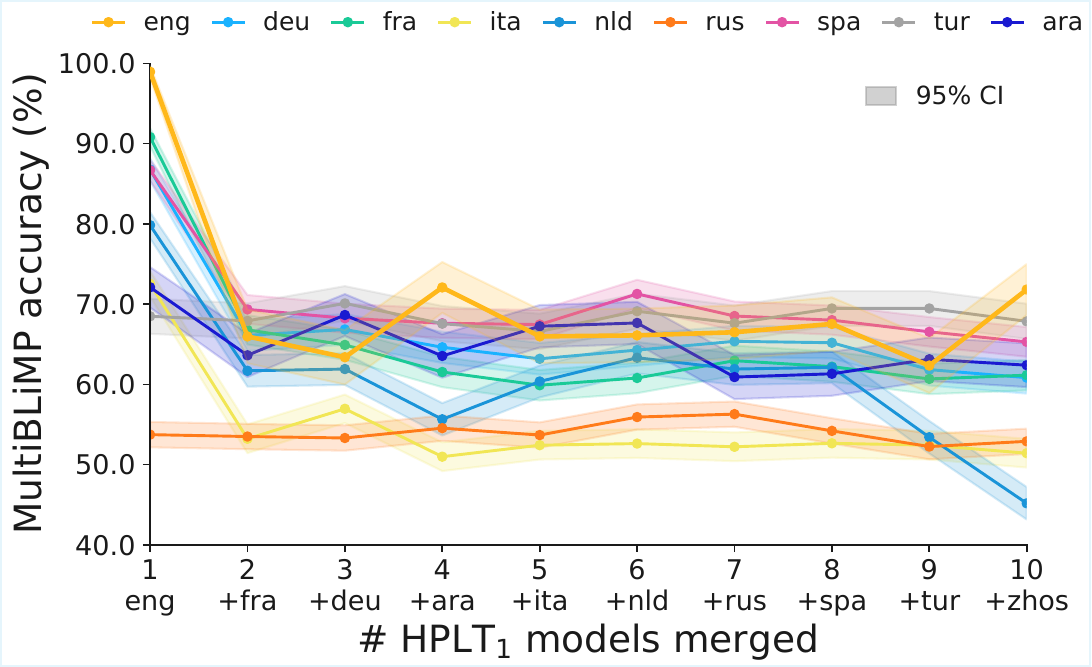}
  \caption{MultiBLiMP accuracy per-language for linear merges of 2-10 monolingual HPLT$_1$ models (randomly ordered; only combination matters). At each step one model is evaluated on all languages. We note HPLT$_{\text{eng}}$ is incidentally performant in various languages. Performance collapses to near-random from the first merge.
  }
  \label{fig:spectrum-merge}
\end{figure}

\paragraph{Merging causes performance collapse} Linear merging of 10 monolingual pre-trained models leads to near-random performance across all benchmarks and languages (see Appendix \ref{app:full-results} for full language results), we expect due to catastrophic interference of conflicting parameters during merging. Figure \ref{fig:spectrum-merge} shows this occurs for any number of merges: merging successive combinations of 2 to 10 models shows MultiBLiMP performance collapses from the first merge, with no clear gains on each added language. Further ablations indicate this result is agnostic to the training stage, as Figure \ref{fig:mix-vs-merge} shows merging equivalent checkpoints throughout training consistently results in near-random performance compared to mixed pre-training.

\paragraph{Interference mitigation does not help} Applying TIES and DARE-TIES merging, which prune unimportant parameters to reduce interference, yields negligible improvements across tasks, with performance still collapsing against monolingual and mixed models. Qualitative analysis of FLORES-200 outputs, seen in \Cref{app:qual-analysis}, suggests \textit{all} merged models fail to generate any meaningful text in any language. This confirms that independently pre-trained models are too distinct in both magnitude and weight-space alignment compared to fine-tuned versions of a pre-trained model, compromising these methods' effectiveness. 

\paragraph{Mixed pre-training gives modest results} Our Mixed$_{10}$ model achieves consistently good performance across languages, underperforming HPLT$_1$ experts but performing similarly to EuroLLM. This suggests mixed pre-training is a good compromise for multilingual settings, while monolingual models are more performant when the task language is pre-defined. This result may arise from both limited token-exposure compared to baselines, and smaller model capacity per-language compared to HPLT$_1$ models. However, Figure \ref{fig:mix-vs-merge} indicates that Mixed$_{10}$ still reaches a high MultiBLiMP accuracy across languages after only 10000 training steps.

\section{Analysis}
\label{sec:analysis}

Our results pose the question of \textit{why} merging harms performance, and to what extent we can predict merge failure or success. We explore this question by calculating performance drops from merging, and testing their correlations with model similarity measures. We linearly merge all bilingual combinations of the 10 HPLT$_1$ models, giving 45 merged models. For each language pair, we calculate $\Delta$ as the mean \textit{drop} in MultiBLiMP accuracy on the two merged languages between the HPLT$_1$ experts and the merged bilingual model. We calculate \textit{parametric} similarity between merged models via layer-wise cosine similarity, mean stable rank difference, and mean L2 norm difference, but find no significant correlations with $\Delta$ (see Appendix \ref{app:correlation}).

\begin{figure}[t!]
  \centering
  \hspace{-0.3cm}
  \includegraphics[width=0.49\textwidth]{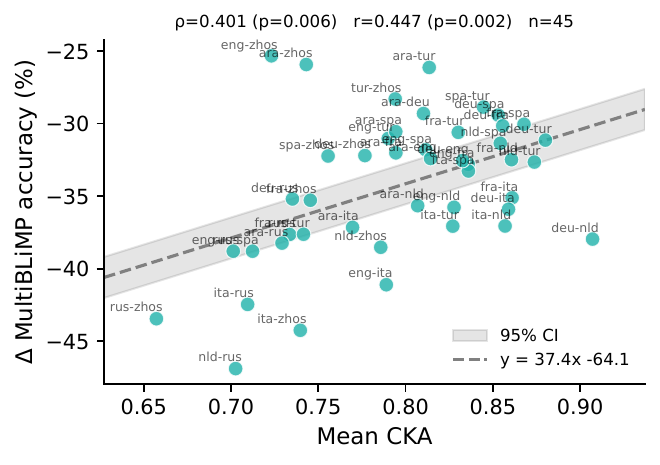}
  \caption{Mean layer-wise CKA between monolingual HPLT$_1$ models significantly correlates (in terms of Spearman's $\rho$ and Pearson's $r$) with smaller merge performance drop $\Delta$ from monolingual to bilingual merged models. This suggests increasing representational similarity improves merge success.}
  \label{fig:sim-correlation}
\end{figure}

We then test \textit{representational} similarity measures, which are invariant to parameter symmetries that can confound weight-space metrics~\citep{Klabunde25-SimilarityNeuralNetwork}. 
We calculate mean layer-wise linear centred kernel alignment (CKA)~\citep{Kornblith19-SimilarityNeuralNetwork} of model representations, by  passing the English FLORES-200 devtest through all 10 HPLT$_1$ models\footnote{We assume all monolingual models have been exposed to non-zero amounts of English text given difficulties in document-level language identification \citep{Fedorova26-OpenLIDv3ImprovingPrecision}.}, then averaging layer-wise CKA. Figure \ref{fig:sim-correlation} shows higher CKA significantly correlates with smaller $\Delta$ ($r=0.447$, $p<0.005$), suggesting cross-model representational similarity reduces interference and is a useful predictor of merge failure.

\section{Discussion}

The inefficacy of merging independently pre-trained monolingual models across all tasks, languages, and merge methods contrasts with the success of merging language-specific fine-tuned models for multilinguality~\citep{Aakanksha24-MixDataMerge, Bandarkar25-UnreasonableEffectivenessModela}.
Our results align with prior work suggesting that representational similarity is an essential ingredient for successful merging of heterogenous models~\citep{shaheen2026is}. This could be achieved through through shared pre-training~\citep{li2022branchtrainmerge}, weight-space alignment methods~\citep{ainsworth2023git}, or adapting unsupervised embedding alignment techniques~\citep{lample2018word}, the latter two providing open directions for future work.

While representational similarity is one ingredient for merging, recent work suggests merging heterogenous models also requires interference mitigation strategies~\citep{Chen26-CanHeterogeneousLanguage}. Our results indicate that the converse holds: interference mitigation alone is not sufficient to overcome the representational divergence of monolingually pre-trained models. While future ablations into the effects of initialisation and partial pre-training may prove enlightening, we speculatively conclude that merging pre-trained models requires both representational similarity and interference mitigation methods.

\section{Conclusion}

We find that merging small, independently pre-trained monolingual models is ineffective, despite applying interference-mitigation strategies. Mixed multilingual data pre-training is a simpler, albeit less flexible, approach to achieve modest but consistent performance across languages. Our results suggest merging requires representational similarity between models, but we find that independently pre-training monolingual models leads to divergent, heterogeneous representations. Therefore, our core recommendation is that representational alignment, most straightforwardly achievable through some initial mixed pre-training, is required for effective language-specific model merging.

\section*{Limitations}
We acknowledge the following limitations of this work. First, we only use models around 2B parameters to maintain strict control over experimental conditions; and we train our mixed model on 100B tokens, which is a smaller scale than state-of-the-art models of similar sizes. Our focus is not on leading performance, and while 100B tokens is enough to see clear trends, further training would strengthen results. Next, we test three standard merging strategies, leaving exploration of stronger alignment-based approaches to future work.  We train only one mixed model as a controlled comparison against the HPLT monolingual models, since any additional pre-training required substantial resources. Similarly, due to resource limitations, it was not feasible to scale results to larger models or bigger datasets. We note recent work explores scaling laws for monolingual and multilingual pre-training~\citep{longpre2026atlas}, and future work could extend our research to explore the scaling properties of merging pre-trained models to understand whether our findings hold for larger models.

\section*{Acknowledgements}
\label{sec:acks}
SA was funded in part by the UvA’s Language Sciences for Social Good project, the City of
Amsterdam, and the Netherlands Organization for Scientific Research (NWO) under project numbers VI.C.192.080 and 2023.017.
SA and AU are grateful to Booking.com, where they first collaborated during SA's internship. This partnership led to the present collaboration, which is unrelated to Booking.com and was carried out independently of the company.
This project has received funding from the Horizon Europe research and innovation programme of the European Union under Grant No.~101070350 and Grant No.~101195233 (Digital Europe programme of the European Union). The authors thank CSC (Finland) for computational resources and support.
We further thank both colleagues from the UvA LTL and Helsinki NLP groups for providing helpful feedback prior to submission, and the anonymous reviewers for their constructive efforts to improve this research.

\bibliography{misc/references.bib, misc/custom.bib}
\bibliographystyle{misc/acl_natbib}

\appendix
\section{Full Task Results}
\label{app:full-results}
We provide full results per task across all available languages for each task in Tables \ref{tab:multiblimp}--\ref{tab:flores200}. We note that findings from Table \ref{tab:main-results} are reflected here: Monolingual expert models perform best on MultiBLiMP, and otherwise multilingual baselines are strong. Mixed pre-training gives consistent but not leading performance across languages, while any merging leads to performance collapse to near-random across all languages.

\begin{table*}[h!]
\centering
\small
\setlength{\tabcolsep}{5pt}

\begin{tabular}{l S[table-format=3.1,table-number-alignment=right] S[table-format=3.1,table-number-alignment=right] S[table-format=3.1,table-number-alignment=right] S[table-format=3.1,table-number-alignment=right] S[table-format=3.1,table-number-alignment=right] S[table-format=3.1,table-number-alignment=right] S[table-format=3.1,table-number-alignment=right] S[table-format=3.1,table-number-alignment=right] S[table-format=3.1,table-number-alignment=right] S[table-format=3.1,table-number-alignment=right]}
\toprule
\textbf{Model} & \textbf{eng} & \textbf{deu} & \textbf{fra} & \textbf{ita} & \textbf{nld} & \textbf{rus} & \textbf{spa} & \textbf{tur} & \textbf{ara} & \textbf{zho} \\
\midrule
\multicolumn{11}{l}{\quad\textit{Multilingual baselines}} \\[2pt]
Tiny-Aya-Base & \multicolumn{1}{r}{\textbf{99.0}} & 99.0 & 99.3 & 97.2 & 96.3 & 97.8 & 98.0 & 92.8 & 95.4 & {\na} \\
EuroLLM-1.7B & 97.8 & 98.0 & 98.8 & 96.4 & 96.9 & 97.2 & 97.5 & 87.8 & 93.0 & {\na} \\
Gemma-2-2B & 98.7 & 97.4 & 97.9 & 93.9 & 94.5 & 95.1 & 96.3 & 89.0 & 90.6 & {\na} \\
\midrule
\multicolumn{11}{l}{\quad\textit{Monolingual experts}} \\[2pt]
HPLT\textsubscript{1} (per-language) & \multicolumn{1}{r}{\textbf{99.0}} & \multicolumn{1}{r}{\textbf{99.1}} & \multicolumn{1}{r}{\textbf{99.4}} & \multicolumn{1}{r}{\textbf{98.3}} & \multicolumn{1}{r}{\textbf{99.1}} & \multicolumn{1}{r}{\textbf{99.2}} & \multicolumn{1}{r}{\textbf{98.4}} & \multicolumn{1}{r}{\textbf{98.7}} & \multicolumn{1}{r}{\textbf{96.9}} & {\na} \\
\midrule
\multicolumn{11}{l}{\quad\textit{Mixed pre-training }} \\[2pt]
Mixed\textsubscript{10} & 97.5 & 99.0 & 99.1 & 97.2 & 98.6 & 97.8 & 97.9 & 95.7 & 95.2 & {\na} \\
\midrule
\multicolumn{11}{l}{\quad\textit{Merged\textsubscript{10} experts}} \\[2pt]
Linear & 71.8 & 60.8 & 61.2 & 51.4 & 45.2 & 52.9 & 65.3 & 67.8 & 62.4 & {\na} \\
TIES & 67.1 & 62.7 & 57.4 & 52.5 & 50.6 & 53.7 & 65.0 & 66.1 & 64.8 & {\na} \\
DARE-TIES & 67.9 & 60.4 & 60.4 & 55.0 & 50.4 & 54.3 & 66.5 & 68.3 & 63.9 & {\na} \\
\bottomrule
\end{tabular}
\caption{MultiBLIMP accuracy ($\uparrow$) for each model and language. Best per-language entry in bold. {\na} = not available.}
\label{tab:multiblimp}
\end{table*}

\begin{table*}[t]
\centering
\small
\setlength{\tabcolsep}{5pt}

\begin{tabular}{l S[table-format=3.1,table-number-alignment=right] S[table-format=3.1,table-number-alignment=right] S[table-format=3.1,table-number-alignment=right] S[table-format=3.1,table-number-alignment=right] S[table-format=3.1,table-number-alignment=right] S[table-format=3.1,table-number-alignment=right] S[table-format=3.1,table-number-alignment=right] S[table-format=3.1,table-number-alignment=right] S[table-format=3.1,table-number-alignment=right] S[table-format=3.1,table-number-alignment=right]}
\toprule
\textbf{Model} & \textbf{eng} & \textbf{deu} & \textbf{fra} & \textbf{ita} & \textbf{nld} & \textbf{rus} & \textbf{spa} & \textbf{tur} & \textbf{ara} & \textbf{zho} \\
\midrule
\multicolumn{11}{l}{\quad\textit{Multilingual baselines}} \\[2pt]
EuroLLM-1.7B & 36.8 & 38.9 & 38.6 & 35.2 & 37.2 & 38.6 & 37.4 & 34.1 & 34.6 & 37.1 \\
Gemma-2-2B & 59.0 & 52.3 & 52.6 & 46.3 & 48.2 & 49.4 & 52.9 & 42.1 & 46.2 & 47.2 \\
Tiny-Aya-Base & \multicolumn{1}{r}{\best{61.0}} & \multicolumn{1}{r}{\best{58.3}} & \multicolumn{1}{r}{\best{58.6}} & \multicolumn{1}{r}{\best{52.0}} & \multicolumn{1}{r}{\best{53.6}} & \multicolumn{1}{r}{\best{54.3}} & \multicolumn{1}{r}{\best{57.4}} & \multicolumn{1}{r}{\best{47.0}} & \multicolumn{1}{r}{\best{57.2}} & \multicolumn{1}{r}{\best{52.9}} \\
\midrule
\multicolumn{11}{l}{\quad\textit{Monolingual experts}} \\[2pt]
HPLT\textsubscript{1} (per-language) & 44.2 & 38.9 & 40.1 & 35.2 & 39.6 & 40.4 & 40.8 & 34.9 & 35.4 & 30.0 \\
\midrule
\multicolumn{11}{l}{\quad\textit{Mixed pre-training }} \\[2pt]
Mixed\textsubscript{10} & 38.3 & 38.8 & 40.8 & 36.3 & 37.9 & 36.6 & 38.4 & 34.3 & 36.1 & 27.2 \\
\midrule
\multicolumn{11}{l}{\quad\textit{Merged\textsubscript{10} experts}} \\[2pt]
Linear & 24.2 & 21.9 & 23.7 & 27.1 & 24.8 & 22.7 & 24.0 & 23.0 & 24.6 & 22.6 \\
TIES & 22.2 & 23.8 & 23.4 & 23.3 & 25.7 & 23.9 & 25.7 & 24.0 & 24.7 & 22.2 \\
DARE-TIES & 24.1 & 24.7 & 23.8 & 23.2 & 24.6 & 23.2 & 25.4 & 24.1 & 24.8 & 22.6 \\
\bottomrule
\end{tabular}
\caption{Belebele accuracy (\(\uparrow\)) per model and language. Best per-language entry in bold.}
\label{tab:belebele}
\end{table*}

\begin{table*}[h!]
\centering
\small
\setlength{\tabcolsep}{5pt}
\label{tab:hellaswag}
\begin{tabular}{l S[table-format=3.1,table-number-alignment=right] S[table-format=3.1,table-number-alignment=right] S[table-format=3.1,table-number-alignment=right] S[table-format=3.1,table-number-alignment=right] S[table-format=3.1,table-number-alignment=right] S[table-format=3.1,table-number-alignment=right] S[table-format=3.1,table-number-alignment=right] S[table-format=3.1,table-number-alignment=right] S[table-format=3.1,table-number-alignment=right] S[table-format=3.1,table-number-alignment=right]}
\toprule
\textbf{Model} & \textbf{eng} & \textbf{deu} & \textbf{fra} & \textbf{ita} & \textbf{nld} & \textbf{rus} & \textbf{spa} & \textbf{tur} & \textbf{ara} & \textbf{zho} \\
\midrule
\multicolumn{11}{l}{\quad\textit{Multilingual baselines}} \\[2pt]
EuroLLM-1.7B & 60.1 & 45.9 & 51.3 & 49.1 & 45.8 & 44.9 & 49.1 & 41.1 & 38.9 & 42.0 \\
Gemma-2-2B & \multicolumn{1}{r}{\best{74.5}} & 50.7 & 58.6 & 53.1 & 50.6 & 51.9 & 58.8 & 45.9 & 40.7 & 50.1 \\
Tiny-Aya-Base & 73.5 & \multicolumn{1}{r}{\best{56.0}} & \multicolumn{1}{r}{\best{61.6}} & \multicolumn{1}{r}{\best{61.8}} & \multicolumn{1}{r}{\best{52.3}} & \multicolumn{1}{r}{\best{55.0}} & \multicolumn{1}{r}{\best{61.2}} & \multicolumn{1}{r}{\best{51.7}} & \multicolumn{1}{r}{\best{50.0}} & \multicolumn{1}{r}{\best{52.5}} \\
\midrule
\multicolumn{11}{l}{\quad\textit{Monolingual experts}} \\[2pt]
HPLT\textsubscript{1} (per-language) & 65.0 & 44.1 & 47.8 & 45.2 & 45.6 & 45.0 & 51.0 & 45.5 & 36.0 & 34.7 \\
\midrule
\multicolumn{11}{l}{\quad\textit{Mixed pre-training }} \\[2pt]
Mixed\textsubscript{10} & 48.5 & 38.4 & 42.3 & 41.7 & 39.8 & 39.2 & 44.4 & 40.0 & 36.0 & 29.4 \\
\midrule
\multicolumn{11}{l}{\quad\textit{Merged\textsubscript{10} experts}} \\[2pt]
Linear & 25.7 & 24.8 & 23.7 & 27.6 & 24.4 & 24.9 & 25.4 & 25.2 & 24.4 & 22.8 \\
TIES & 26.3 & 26.2 & 21.5 & 25.8 & 24.0 & 26.4 & 26.1 & 26.3 & 24.9 & 23.2 \\
DARE-TIES & 27.1 & 26.6 & 25.0 & 26.2 & 24.4 & 25.0 & 26.5 & 25.7 & 24.6 & 23.2 \\
\bottomrule
\end{tabular}
\caption{HellaSwag accuracy normalised (\(\uparrow\)) per model and language. Best per-language entry in bold.}
\end{table*}

\begin{table*}[t]
\centering
\small
\setlength{\tabcolsep}{5pt}
\label{tab:xcsqa}
\begin{tabular}{l S[table-format=3.1,table-number-alignment=right] S[table-format=3.1,table-number-alignment=right] S[table-format=3.1,table-number-alignment=right] S[table-format=3.1,table-number-alignment=right] S[table-format=3.1,table-number-alignment=right] S[table-format=3.1,table-number-alignment=right] S[table-format=3.1,table-number-alignment=right] S[table-format=3.1,table-number-alignment=right] S[table-format=3.1,table-number-alignment=right] S[table-format=3.1,table-number-alignment=right]}
\toprule
\textbf{Model} & \textbf{eng} & \textbf{deu} & \textbf{fra} & \textbf{ita} & \textbf{nld} & \textbf{rus} & \textbf{spa} & \textbf{tur} & \textbf{ara} & \textbf{zho} \\
\midrule
\multicolumn{11}{l}{\quad\textit{Multilingual baselines}} \\[2pt]
EuroLLM-1.7B & 42.6 & 34.4 & 29.1 & 32.9 & 33.7 & 25.4 & 29.0 & {\na} & 27.6 & 35.3 \\
Gemma-2-2B & \multicolumn{1}{r}{\best{68.2}} & 48.6 & 43.9 & 42.9 & 38.0 & 33.2 & 45.2 & {\na} & 36.8 & 45.9 \\
Tiny-Aya-Base & 65.1 & \multicolumn{1}{r}{\best{49.9}} & \multicolumn{1}{r}{\best{47.7}} & \multicolumn{1}{r}{\best{51.1}} & \multicolumn{1}{r}{\best{45.5}} & \multicolumn{1}{r}{\best{38.8}} & \multicolumn{1}{r}{\best{48.8}} & {\na} & \multicolumn{1}{r}{\best{43.3}} & \multicolumn{1}{r}{\best{47.6}} \\
\midrule
\multicolumn{11}{l}{\quad\textit{Monolingual experts}} \\[2pt]
HPLT\textsubscript{1} (per-language) & 54.4 & 39.1 & 35.5 & 36.4 & 37.9 & 31.7 & 38.1 & {\na} & 31.2 & 34.8 \\
\midrule
\multicolumn{11}{l}{\quad\textit{Mixed pre-training }} \\[2pt]
Mixed\textsubscript{10} & 45.7 & 35.8 & 33.7 & 37.0 & 34.7 & 30.6 & 35.2 & {\na} & 30.9 & 24.3 \\
\midrule
\multicolumn{11}{l}{\quad\textit{Merged\textsubscript{10} experts}} \\[2pt]
Linear & 21.5 & 20.1 & 18.3 & 22.5 & 20.4 & 22.1 & 21.3 & {\na} & 19.1 & 19.6 \\
TIES & 21.9 & 20.1 & 19.8 & 22.8 & 21.1 & 20.2 & 20.4 & {\na} & 19.3 & 20.3 \\
DARE-TIES & 21.3 & 20.4 & 19.2 & 21.7 & 20.7 & 21.8 & 22.0 & {\na} & 19.2 & 18.3 \\
\bottomrule
\end{tabular}
\caption{X-CSQA accuracy normalised (\(\uparrow\)) per model and language. Best per-language entry in bold. {\na} = not available.}
\end{table*}

\begin{table*}[t]
\centering
\small
\setlength{\tabcolsep}{5pt}

\begin{tabular}{l S[table-format=3.1,table-number-alignment=right] S[table-format=3.1,table-number-alignment=right] S[table-format=3.1,table-number-alignment=right] S[table-format=3.1,table-number-alignment=right] S[table-format=3.1,table-number-alignment=right] S[table-format=3.1,table-number-alignment=right] S[table-format=3.1,table-number-alignment=right] S[table-format=3.1,table-number-alignment=right] S[table-format=3.1,table-number-alignment=right] S[table-format=3.1,table-number-alignment=right]}
\toprule
\textbf{Model} & \textbf{eng} & \textbf{deu} & \textbf{fra} & \textbf{ita} & \textbf{nld} & \textbf{rus} & \textbf{spa} & \textbf{tur} & \textbf{ara} & \textbf{zho} \\
\midrule
\multicolumn{11}{l}{\quad\textit{Multilingual baselines}} \\[2pt]
EuroLLM-1.7B & 46.3 & 41.2 & \multicolumn{1}{r}{\best{59.0}} & \multicolumn{1}{r}{\best{43.3}} & 35.4 & 34.0 & 33.9 & \multicolumn{1}{r}{\best{48.0}} & 27.6 & 8.0 \\
Gemma-2-2B & 43.9 & \multicolumn{1}{r}{\best{52.0}} & 53.7 & 33.5 & \multicolumn{1}{r}{\best{42.7}} & 47.6 & \multicolumn{1}{r}{\best{47.5}} & 38.8 & 29.0 & \multicolumn{1}{r}{\best{13.5}} \\
Tiny-Aya-Base & \multicolumn{1}{r}{\best{51.4}} & 44.2 & 56.1 & 26.4 & 37.0 & \multicolumn{1}{r}{\best{50.6}} & 33.3 & 41.2 & \multicolumn{1}{r}{\best{40.2}} & 7.1 \\
\midrule
\multicolumn{11}{l}{\quad\textit{Monolingual experts}} \\[2pt]
HPLT\textsubscript{1} (per-language) & 34.9 & 27.7 & 55.8 & 42.6 & 32.8 & 26.1 & 43.7 & 35.1 & 23.8 & 5.5 \\
\midrule
\multicolumn{11}{l}{\quad\textit{Mixed pre-training }} \\[2pt]
Mixed\textsubscript{10} & 42.2 & 38.5 & 40.1 & 34.5 & 26.7 & 29.7 & 35.6 & 35.0 & 30.3 & 3.5 \\
\midrule
\multicolumn{11}{l}{\quad\textit{Merged\textsubscript{10} experts}} \\[2pt]
Linear & 2.1 & 1.8 & 3.0 & 1.9 & 2.0 & 2.1 & 2.3 & 1.9 & 1.9 & 2.4 \\
TIES & 15.8 & 15.9 & 16.4 & 17.7 & 16.7 & 4.8 & 16.7 & 14.1 & 3.9 & 4.9 \\
DARE-TIES & 1.5 & 2.2 & 1.9 & 2.3 & 1.4 & 1.9 & 0.9 & 1.6 & 1.6 & 2.3 \\
\bottomrule
\end{tabular}
\caption{FLORES-200 ChrF++ (eng--xxx, \(\uparrow\)) per model and language. eng results are for fra--eng. Best per-language entry in bold.}
\label{tab:flores200}
\end{table*}

\section{Qualitative Analysis of Outputs}
\label{app:qual-analysis}

 We perform a brief qualitative analysis of model generations for the FLORES translation task, with examples shown in Table \ref{tab:qual-examples}. We see the following error modes: monolingual models show input copying and occasionally fail to translate into the target language. Some models including Tiny-Aya fail to stop generating after translating. The Mixed$_{10}$ model generally provides a functional translation but sometimes still copies inputs. Finally, all merging settings lead to complete output collapse: either generating strings of numerals, or nonsense tokens. This suggests any merging leads models to completely lose their generative language capabilities. 

\begin{table*}[t]
  \centering
  \small
  \begin{tabular}{lp{13cm}}
    \toprule
    \textbf{Model} & \textbf{Translation} \\
    \midrule
    \textsc{Source} & Del Potro had the early advantage in the second set, but this too required a tie break after reaching 6-6. \\
    \textsc{Target} & Malgré le net avantage de Del Potro pendant le deuxième set, il a fallu passer par un tie-break une fois que le score a atteint 6-6. \\
    \midrule
    \textsc{Tiny-Aya} & Del Potro a eu un avantage précoce dans le deuxième set, mais cela aussi a nécessité un tie break après avoir atteint 6-6. \\
HPLT$_{\text{eng}}$ & Del Potro had the early advantage in the second set, but this too required a tie break after reaching 6-6. \\
HPLT$_{\text{fra}}$ & L'Argentin Del Potro avait l'avantage au début de la deuxième manche, mais cette fois-ci, il a fallu un tie break après avoir atteint 6-6. \\
\textsc{Mixed$_{10}$} & Del Potro avait l'avantage dans le premier set, mais ce fut un tie-break après avoir atteint 6-6. \\
\textsc{Merged$_{10}$} & "0000000000000000000000000000000000000000000000000000000000000000000000000000000\ldots{} \\
\textsc{Ties$_{10}$} & y velalixobalixobalix ...,obal:'al:'al:'al.........gerixyinternetal.........geryinternetal.........gerix ...,obal.........gerstonal.........yideaideaideaideaideaideaideaideaideaideaideaideaideaideaide\ldots{} \\
\textsc{Dare-Ties$_{10}$} & 01999999999999999999999999999999999999999999999999999999999999999999999999999999\ldots{} \\
    \midrule
    \midrule
    
        \textsc{Source} & The 35mm format is actually, somewhat confusingly, 36mm in width by 24mm in height. \\
    \textsc{Target} & Le format 35 mm est en fait, et c’est assez déconcertant, de 36 mm de largeur par 24 mm de hauteur. \\
    \midrule
    \textsc{Tiny-Aya} & Le format 35 mm est en fait, de manière quelque peu déroutante, de 36 mm de large par 24 mm de haut. \\
HPLT$_{\text{eng}}$ & Le format 35mm n’est pas vraiment confusingement confusingement 36mm en hauteur par 24mm en largeur. \\
HPLT$_{\text{fra}}$ & Le format 35mm est en fait, quelque peu confus, 36mm en largeur par 24mm en hauteur. \\

    \textsc{Mixed$_{10}$} & The 35mm format is actually, somewhat confusingly, 36mm in width by 24mm in height. \\
\textsc{Merged$_{10}$} & 00000000000000000000000000000000000000000000000000000000000000000000000000000000\ldots{} \\  
    \textsc{Ties$_{10}$} & yyyyyinternetalral...)alral...)al.........geryinternetal.........gerixyinternetal.........geryinternetal......... gerixyinternetal.........gerstoneritisnetal.........ireoveoveoveoveoveoveoveoveoveoveov\ldots{} \\
\textsc{Dare-Ties$_{10}$} & 00000000000000000000000000000000000000000000000000000000000000000000000000000000\ldots{} \\
    \midrule
    \midrule

        \textsc{Source} & It is related to but usually not involving alpine style ski touring or mountaineering, the latter ones done in steep terrain and requiring much stiffer skis and boots. \\
    \textsc{Target} & Cela ressemble au ski de randonnée de type alpin, ainsi qu'à l'alpinisme, sans toutefois incorporer en général ces activités, qui se pratiquent sur des terrains accidentés et qui exigent l'utilisation\ldots{} \\
    \midrule
    \textsc{Tiny-Aya} & C'est lié à mais généralement pas impliquant le ski-touring ou l'alpinisme de style alpin, ces derniers étant pratiqués dans des terrains raides et nécessitant des skis et des bottes beaucoup plus rigides.\\ & English: It is related to but usually not involving alpine style ski touring or mountaineering, the latter ones done \\
 HPLT$_{\text{eng}}$ & It is related to but usually not involving alpine style ski touring or mountaineering, the latter ones done in steep terrain and requiring much stiffer skis and boots. \\
 HPLT$_{\text{fra}}$ & It is related to but usually not involving alpine style ski touring or mountaineering, the latter ones done in steep terrain and requiring much stiffer skis and boots.\\& Le ski de randonnée est une activité de plein air qui consiste à parcourir des distances de plus en plus longues à l'aide de skis spécialement conçus à cet effet.\\
\textsc{Mixed$_{10}$} & It is related to but usually not involving alpine style ski touring or mountaineering, the latter ones done in steep terrain and requiring much stiffer skis and boots. \\
\textsc{Merged$_{10}$} & "0000000000000000000000000000000000000000000000000000000000000000000000000000000\ldots{} \\
\textsc{Ties$_{10}$} & dalaodyinternetiiobalix ...,obalix ...,obal.........yinternetal .........yideaideaideaideaideaideaideaidea ideaideaideaidea- ...al.........yodononet ' ' ' ' ' ' ' ' ' ' ' ' ' ' ' ' ' ' ' ' ' ' ' ' ' ' ' \ldots{} \\
\textsc{Dare-Ties$_{10}$} & 88888899999999999999999999999999999999999999999999999999999999999999999999999999\ldots{} \\

    \bottomrule
  \end{tabular}
  \caption{Example generated translations from various models (FLORES devtest, English$\to$French). All merging leads to catastrophic collapse of generation capabilities, while monolingual models exhibit input copying and/or a failure to translate.}
  \label{tab:qual-examples}
\end{table*}

\section{Model Similarity Correlation}

\label{app:correlation}
In Table \ref{tab:correlations}, we report results of Pearson's $r$ and Spearman's $\rho$ correlation tests between $\Delta$ performance after merging and measures of model and language similarity: layer-wise cosine similarity, mean layer-wise rank difference, absolute mean L2 norm difference across layers, mean layer-wise CKA~\citep{Kornblith19-SimilarityNeuralNetwork}, and Lang2vec typological distance (with kNN imputation of missing features from similar languages) of the 2 merged languages~\citep{littell-etal-2017-uriel, goot-etal-2025-distals}.
\begin{table}[H]
\centering
\small
\begin{tabular}{@{}lrrrr@{}}
\toprule
Measure & $\rho$ & $p_S$ & $r$ & $p_P$ \\
\midrule
Mean CKA & 0.40$^{**}$ & 0.006 & 0.45$^{**}$ & 0.002 \\
Lang2vec (kNN) & 0.35$^{*}$ & 0.017 & 0.32$^{*}$ & 0.033 \\
Mean Rank $\Delta$ & 0.15 & 0.340 & 0.11 & 0.458 \\
Cosine Similarity & 0.11 & 0.468 & 0.11 & 0.460 \\
Mean L2 Norm $\Delta$ & 0.03 & 0.830 & 0.04 & 0.798 \\
\bottomrule
\end{tabular}
\caption{Spearman and Pearson correlations of various model- and language-similarity measures against $\Delta$ MultiBLiMP accuracy for 45 bilingual merges of monolingual models. $^{*}p<0.05$, $^{**}p<0.01$.}
\label{tab:correlations}
\end{table}

We observe no significant correlation between parametric similarity measures such as layer-wise cosine similarity. However, we observe a surprising correlation between \textit{increasing} Lang2vec typological distance and \textit{decreasing} $\Delta$, i.e. smaller performance drops. This indicates that combining more typologically similar languages results in worse performing merges. We leave exploration of this to future work.

\end{document}